\documentclass[conference]{IEEEtran}
\IEEEoverridecommandlockouts
\usepackage{cite}
\usepackage{amsmath,amssymb,amsfonts}
\usepackage{algorithmic}
\usepackage{graphicx}
\usepackage{textcomp}
\usepackage{xcolor}
\usepackage{makecell}
\usepackage{multirow}

\usepackage{bbding}
\usepackage{color}
\usepackage{subfigure}
\usepackage{times}

\usepackage{soul}
\usepackage[hidelinks]{hyperref}
\usepackage[utf8]{inputenc}
\usepackage[small]{caption}
\usepackage{graphicx}
\usepackage{amsmath}
\usepackage{booktabs}
\usepackage{mathrsfs}
\usepackage{amsfonts,amssymb}
\usepackage{multirow}
\usepackage{color}
\usepackage{cite}
\usepackage{pifont}
\usepackage{array}
\usepackage[misc]{ifsym} 
\usepackage{makecell}
\usepackage{mathrsfs}
\usepackage{epstopdf}
\usepackage{threeparttable}

\usepackage{CJKutf8}
\usepackage{geometry}

\def\BibTeX{{\rm B\kern-.05em{\sc i\kern-.025em b}\kern-.08em
T\kern-.1667em\lower.7ex\hbox{E}\kern-.125emX}}
\begin{document}
\title{From Prompt Optimization to Multi-Dimensional Credibility Evaluation: Enhancing Trustworthiness of Chinese LLM-Generated Liver MRI Reports - with Preliminary Extension to Lung Cancer
\\
\thanks{
Corresponding authors: X.~Li and W.~Chen. Q.~Wang, X. Sun contributed equally to this manuscript.}
}

\author{
    \IEEEauthorblockN{
        Qiuli~Wang\textsuperscript{1,2},
        Xinhuan~Sun\textsuperscript{1},
        Yonglin~Chen\textsuperscript{2},
        Jie~Cheng\textsuperscript{2},
        Yongxu~Liu\textsuperscript{1},
        Xingpeng~Zhang\textsuperscript{3},\\
        Xiaoming~Li\textsuperscript{2},
        and Wei~Chen\textsuperscript{2}}

    \IEEEauthorblockA{\textsuperscript{1}Yu-Yue Pathology Research Center, Jinfeng Laboratory, Chongqing, China\\}
    \IEEEauthorblockA{\textsuperscript{2}7T Magnetic Resonance Imaging Translational Medical Center, Department of Radiology,\\ Southwest Hospital, Army Medical University, Chongqing, China\\}   
    \IEEEauthorblockA{\textsuperscript{3}School of Computer Science and Software Engineering, \\Southwest Petroleum University, Chengdu, China\\}
    }

\maketitle

\begin{abstract}
  \emph{Background:}
  Large language models (LLMs) have demonstrated promising performance in generating 
  diagnostic conclusions from imaging findings, thereby supporting radiology reporting, 
  trainee education, and quality control. However, systematic guidance on how to optimize 
  prompt design across different clinical contexts remains underexplored. Moreover, a 
  comprehensive and standardized framework for assessing the trustworthiness of 
  LLM-generated radiology reports is yet to be established.
  
  \emph{Purpose: }
  This study aims to enhance the trustworthiness of LLM-generated liver MRI reports 
  by introducing a Multi-Dimensional Credibility Assessment (MDCA) framework and 
  providing guidance on institution-specific prompt optimization. The proposed framework 
  is applied to evaluate and compare the performance of several advanced LLMs using the SiliconFlow platform. Cross-center 
  generalizability and cross-disease applicability were further examined using an 
  external liver MRI cohort and an in-house lung cancer dataset.
  
  \emph{Materials and Methods:}
  Four large language models (Kimi-K2, Qwen3-235B,
  DeepSeek-V3, and ByteDance-Seed) were used to generate diagnostic
  conclusions from liver MRI findings via the SiliconFlow platform. Report quality
  was assessed using the Multi-Dimensional Credibility Assessment (MDCA) framework
  across three dimensions: Semantic Coherence (SC), Diagnostic Correctness (DC), and
  Clinical Prioritization Alignment (CPA). Eleven prompt configurations were compared,
  combining role definition, task specification, tiered diagnostic taxonomy,
  verification items, structure standards, imaging principles, and example-based
  guidance. Validation was conducted on 15,127 liver MRI reports from the primary
  center, with 1,009 external reports and 1692 lung cancer reports used to test
  cross-center and cross-disease generalizability.

\emph{Results:}
  The MDCA framework effectively distinguished report quality across prompts and
  models. Optimized prompts markedly improved diagnostic correctness and clinical
  prioritization, whereas example-based prompts primarily enhanced semantic coherence.
  Under limited computational resources, approximately ten examples provided the
  optimal performance-efficiency trade-off. Kimi-K2 achieved the highest
  composite score (0.7615) (2025/11), followed closely by DeepSeek-V3 (0.7541). The optimized
  prompt strategy yielded consistent gains on both external liver MRI and lung cancer
  datasets, confirming cross-center and cross-disease generalizability.

\emph{Conclusion:}
  Institution-specific prompt optimization paired with multi-dimensional credibility
  assessment substantially improves the trustworthiness of LLM-generated liver MRI
  reports, with cross-center and cross-disease validation confirming generalizability.
  These approaches enhance reliability and interpretability while providing practical
  tools for radiology quality control and trainee education, supporting the safe
  integration of LLMs into clinical workflows.

  \end{abstract}
  

\IEEEpeerreviewmaketitle
\section{Introduction}
\IEEEPARstart{H}{epatic} space-occupying lesions, particularly hepatocellular carcinoma (HCC), are highly prevalent in China and pose a substantial clinical challenge \cite{llovet2022immunotherapies,bruix2005management,VOGEL20221345}. Gadoxetate disodium (Gd-EOB-DTPA)-enhanced MRI has become an indispensable modality for lesion characterization, early detection, and surgical planning \cite{aoki2021higher,ye2024gd}. At the Department of Radiology, The First Affiliated Hospital of Army Medical University in Chongqing, China, the steadily increasing demand for liver-specific MRI examinations has placed mounting pressure on radiologists. The sheer volume and complexity of imaging data frequently compromise report quality, resulting in vague terminology, inconsistent interpretations, and suboptimal standardization.

The rapid advancement of Chinese large language models (LLMs)—including Kimi-K2, Qwen3, DeepSeek-V3, and ByteDance-Seed—has opened new avenues for radiology report generation in native-language clinical practice \cite{wu2025large,team2025kimi,yang2025qwen3,seed2025seed1,chen2025fuzzy}. These models demonstrate strong capabilities in producing coherent, standardized, and stylistically consistent diagnostic narratives, while also supporting quality control and radiologist training \cite{gaber2025evaluating,marrocchio2025will,sandmann2025benchmark,tordjman2025comparative}. However, their performance in liver MRI reporting remains heterogeneous across tasks and institutions, leaving two major challenges unresolved:

\emph{(1) How can the credibility of LLM-generated radiology reports be objectively evaluated?}
Existing approaches often rely on LLMs themselves for output quality assessment, which incurs substantial computational overhead, constrains real-time deployment, and requires additional evaluation prompts. Moreover, such self-assessment methods risk perpetuating model-intrinsic biases and lack interpretability \cite{doi:10.1148/rg.2017170047,wang2025credibility}. Although expert validation by radiologists could improve reliability, manually reviewing tens of thousands of reports is practically infeasible given the prohibitive workload and time demands. To date, no standardized, objective framework exists for assessing the semantic integrity and diagnostic reliability of LLM-generated reports in a model-agnostic and reproducible manner \cite{tu2025towards}.

\emph{(2) How can LLMs be effectively guided to interpret imaging findings and formulate diagnostic conclusions through prompt design?}
Although several studies have explored various prompting strategies, most remain narrowly tailored to specific tasks or datasets and lack adaptability to institutional reporting conventions. This specificity limits generalizability and impedes clinical deployment across heterogeneous environments. In the absence of well-established design principles, prompt engineering remains largely empirical, driven by iterative trial-and-error rather than systematic guidance \cite{kim2025optimizing,fink2023potential}.

To address these challenges, we developed a unified prompt design protocol compatible with multiple LLMs and established an objective evaluation framework to assess the credibility and quality of LLM-generated radiology reports. First, we systematically evaluated 11 customized prompting strategies—ranging from minimal to example-based and checklist-guided designs—to assess their influence on report reliability and consistency. These strategies were validated on a primary-center liver MRI cohort (n = 15,127) and further examined on an external-center cohort (n = 1,009) to confirm cross-center generalizability. Second, we introduced a Multi-Dimensional Credibility Assessment (MDCA) framework that evaluates report quality across three clinically relevant dimensions: Semantic Coherence (SC), Diagnostic Correctness (DC), and Clinical Prioritization Alignment (CPA) \cite{zhou2021visual,chen2025visual}. Finally, applying the MDCA framework, we benchmarked multiple state-of-the-art Chinese LLMs—including Kimi-K2-Instruct-0905, Qwen3-235B-A22B-Instruct-2507, DeepSeek-V3, and ByteDance-Seed-OSS-36B-Instruct—on the same SiliconFlow platform to identify practical strategies for improving reporting quality, supporting radiology quality assurance, and facilitating trainee education \cite{team2025kimi,seed2025seed1,sandmann2025benchmark,chou2025autocodebench}. Cross-disease applicability was further assessed using 1692 lung cancer reports from the primary center.

\section{Materials and Methods}
This retrospective study was conducted at Department of Radiology, The First Affiliated Hospital of Army Medical University. The requirement for informed consent was waived due to the use of de-identified patient data. Liver radiology reports without diagnostic conclusions were input to the LLMs between April 30, 2025, and May 14, 2025.

\subsection{Collection of Clinical Reports}
\begin{figure}[htbp]
  \centerline{\includegraphics[width=0.5\textwidth]{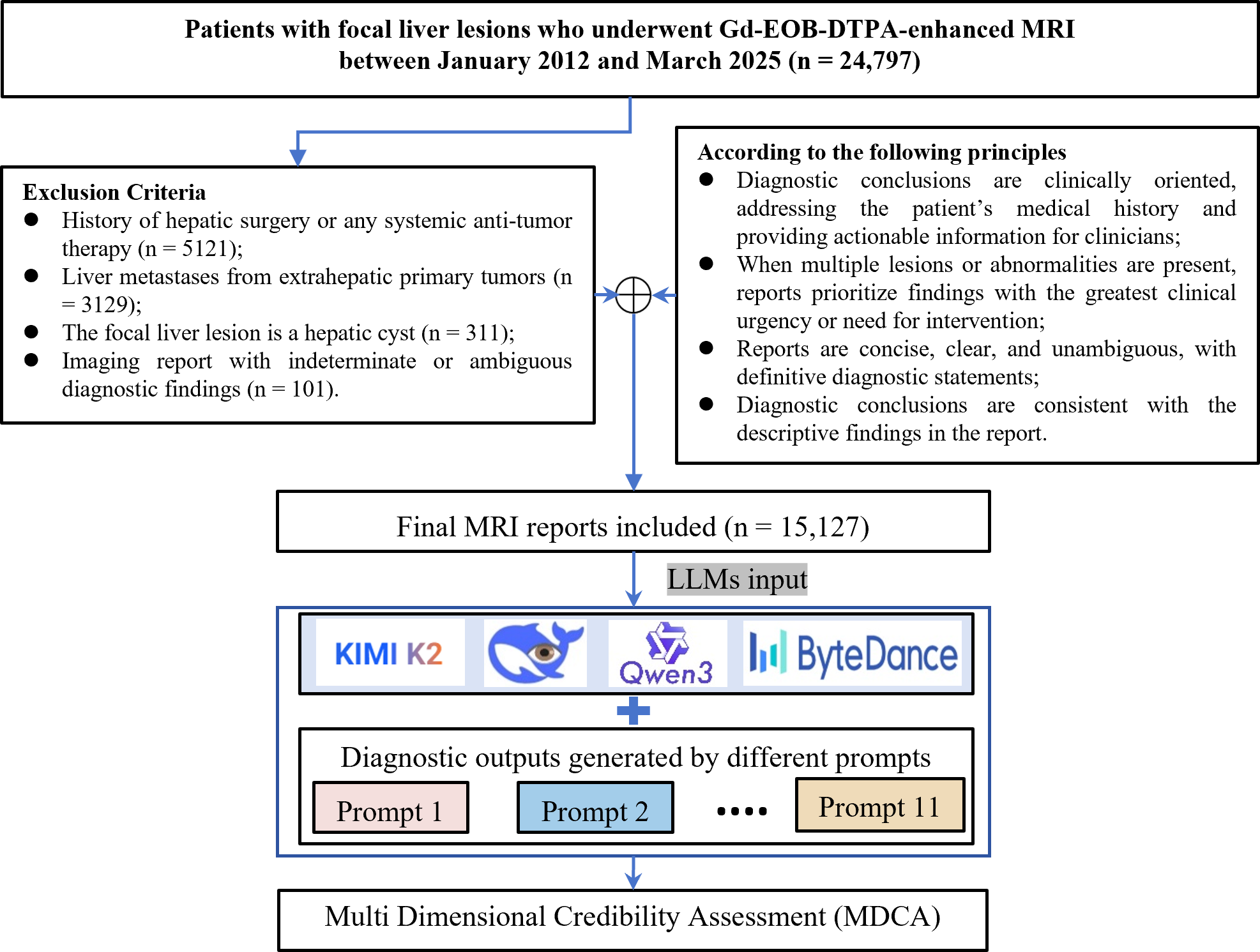}}
  \caption{The collection process of clinical reports. This process followed a well-designed exclusion criteria to ensure the quality and relevance of the reports used for LLM evaluation.}
  \label{fig1}
  \end{figure}

The collection process of clinical reports is shown in Figure~\ref{fig1}. All reports were obtained from our center for patients who underwent Gd-EOB-DTPA-enhanced liver MRI between January and December 2024 (n = 24,797). Exclusion criteria included a history of hepatic surgery or any systemic anti-tumor therapy, a history of extrahepatic primary malignancy, absence of any benign or malignant hepatic lesions, and poor-quality MRI images that prevented reliable assessment. The review process yielded 15,127 eligible liver MRI reports for inclusion in the study.

\subsection{Multi-Dimensional Credibility Assessment Framework}
To independently evaluate the synthesized reports, we proposed a MDCA framework, which evaluated the synthesized reports in three dimensions: SC, DC, and CPA \cite{zhou2021visual,chen2025visual,zeng2025deepseek,bhayana2024large}. 

\emph{Semantic Coherence (SC)}: This metric assessed the overall semantic alignment between the generated diagnostic conclusion and the reference diagnostic conclusion 
at the document level, indicating how closely the LLM output adhered to the linguistic style and radiological writing conventions of the target institution. Specifically, a pre-trained BERT model was used to encode the complete diagnostic conclusion text of the target report $R_T$ and that of the synthesized report $R_S$ into fixed-length vector representations, following the semantic textual similarity (STS) paradigm. The SC score was defined as the cosine similarity between the two document-level embeddings:
\begin{align}
  SC = \cos\!\big(\mathrm{BERT}(R_T),\; \mathrm{BERT}(R_S)\big)
\end{align}

where $\mathrm{BERT}(\cdot)$ denotes the pooled output embedding of the full diagnostic conclusion. Higher SC values reflected greater semantic similarity to the reference report, indicating better fluency and stronger alignment with institutional radiological writing conventions. This metric focused solely on textual and stylistic coherence and did not assess diagnostic accuracy.

\emph{Diagnostic Correctness (DC)}: This dimension evaluated the diagnostic 
accuracy and completeness of the synthesized report relative to the target 
report. The assessment followed a two-stage keyword-level matching procedure, 
termed KMAT. In the extraction stage, diagnostic keywords were identified 
from both the target report $R_T$ and the synthesized report $R_S$ against 
a predefined institutional lexicon of medical diagnostic terms, yielding 
keyword sets $\mathcal{K}_T = \{k_{T,1}, k_{T,2}, \dots, k_{T,m}\}$ and 
$\mathcal{K}_S = \{k_{S,1}, k_{S,2}, \dots, k_{S,n}\}$, respectively. 
In the matching stage, each keyword in $\mathcal{K}_T$ was checked against 
all keywords in $\mathcal{K}_S$ to determine whether the corresponding 
diagnostic entity had been correctly captured. The DC score was then defined 
as the proportion of target keywords that were successfully matched:

\begin{align}
  DC = &\text{KMAT}(R_T, R_S) 
     \\= &\frac{|\{\,k \in \mathcal{K}_T : \exists\, k' \in \mathcal{K}_S,\; 
        \text{match}(k, k')\,\}|}
            {|\mathcal{K}_T|}
\end{align}

A higher DC score indicated stronger diagnostic coverage and terminology 
precision, reflecting that the synthesized report captured a greater 
proportion of clinically relevant findings \cite{savage2025radsearch,
yasaka2025new,lucas2025multisequence}.

\emph{Clinical Prioritization Alignment (CPA)}: This metric assessed whether the most clinically urgent or important findings were presented at the top of the report, consistent with real-world diagnostic priorities. For example, malignant tumors or acute conditions (e.g., hemorrhage) should have appeared at the very first. Scoring was based on whether such high-priority diagnoses appeared within the top 1, top 3, or top 5 positions in the report. Expert radiologists validated prioritization alignment according to institutional guidelines. Based on the clinical requirements in the Department of Radiology, The First Affiliated Hospital of Army Medical University (Southwest Hospital) , the calculation process was defined in two steps:

Step 1: $t_{RT1}$ would be compared with $t_{ST1}$. If they match, a score of 1 was assigned; otherwise, the score was 0. Thus, the score for this step $X_1$ was either 1 or 0.;

Step2: each $t_{RTo}$ would be compared with $t_{STo-1}$, $t_{STo}$, and $t_{STo+1}$, where $o>1$, was the sentence number. For example, $t_{RT2}$ would be matched with $t_{RS1}$, $t_{RS2}$, and $t_{RS3}$. This pattern continued accordingly. Each matched diagnosis was assigned a score of 1, and the total score was obtained by summing all matched instances. The total score for given reports was then normalized by dividing the number of matched diagnoses by the total number of candidate diagnoses, resulting in a final score $X_2$  between 0 and 1.
The final CPA was calculated as:
\begin{align}
  CPA=0.5\times X_1+0.5\times X_2
\end{align}

In addition, Top-1 matching scores were calculated to further evaluate model performance in identifying the key diagnostic conclusions. A score of 1 was assigned when the model's primary diagnosis exactly matched the ground truth, and 0 otherwise.
MDCA Score: The final MDCA score was defined as follow:
\begin{align}
  MDCA =0.4\times SC+0.2\times DC+0.4\times CPA
\end{align}

Generally, all three dimensions contribute to the overall credibility of a 
generated report, but their clinical priorities differ. In this framework, 
SC and CPA were each assigned a weight of 0.4, reflecting the importance of 
both linguistic conformity to institutional standards and clinically prioritized 
content organization. DC, which assesses the accuracy of individual diagnostic 
terms, was assigned a lower weight of 0.2. This is because diagnostic correctness 
is highly dependent on the completeness of the finding description and the 
specificity of the diagnostic lexicon, and minor terminological mismatches 
do not necessarily indicate clinically meaningful errors. The weight assignment 
is adjustable and can be adapted to the requirements of different institutions.

\subsection{Prompt Design and Implementation}
As shown in Table~\ref{prompt_summary}, the prompt content were categorized into two major types: instruction-based components, and example-based guidance. The instruction-based components contained (1) role specification, (2) task definition, (3) tiered diagnostic taxonomy (TOP system), (4) mandatory verification checkpoints, (5) report formatting standards, and (6) radiologic diagnostic principles. The example-based component consisted of sample reports written by experienced radiologists, which provided reference style and diagnostic reasoning patterns. 

\begin{CJK}{UTF8}{gbsn}
\begin{table*}[htbp]
\centering
\caption{Summary of Prompt Design Components and Example Content}
\renewcommand{\arraystretch}{1.4}
\begin{tabular}{@{}p{3cm}p{3cm}p{7cm}@{}}
\toprule
\textbf{Component (EN)} & \textbf{Component (中文)} & \textbf{Example Content (EN / 中文)} \\
\midrule
Role Specification & 角色设定 & You are a radiologist with 30 years of experience in liver MRI interpretation, specializing in HCC and FNH ... \newline 你是一名具有30年肝脏MRI诊断经验的腹部影像诊断医师，专长良恶性肝占位鉴别（尤其FNH/血管瘤/ICC/HCC）... \\
\midrule
Task Definition & 基本任务要求 & Prioritize malignant tumors and avoid speculative conclusions. Follow diagnostic standards ... \newline 强制优先诊断恶性肿瘤；禁止推测性结论，严格遵循诊断标准 ... \\
\midrule
Tiered Diagnostic Taxonomy (TOP System) & 分级诊断体系（TOP） & \textbf{TOP1}: HCC, ICC, metastases,... \newline \textbf{TOP2}: FNH, hemangioma, ... \newline \textbf{TOP3}: Cirrhosis, portal hypertension, ... \newline \textbf{TOP4}: Lymphadenopathy, biliary stones, ... \newline \textbf{TOP5}: Renal cysts, pleural effusion, ... \newline --- \newline \textbf{TOP1}：HCC、胆管细胞癌、转移瘤等， ... \newline \textbf{TOP2}：局灶性结节增生，血管瘤等 \newline \textbf{TOP3}：肝硬化、门脉高压等 \newline \textbf{TOP4}：淋巴结增大、胆系结石等 \newline \textbf{TOP5}：肾囊肿、胸腔积液等 \\
\midrule
Mandatory Verification Checkpoints & 强制核查项 & Confirm HCC, metastases, perfusion anomalies, cirrhosis, lymphadenopathy, renal cysts, ... \newline ☑ 肝癌 ☑ 转移灶 ☑ 灌注异常 ☑ 肝硬化 ☑ 淋巴结 ☑ 肾囊肿等，每项需逐一确认。 \\
\midrule
Report Formatting Standards & 报告结构规范 & Format: anatomical location + disease, one line per diagnosis, ordered by TOP level ... \newline 格式：解剖部位+病变类型，每行为一条，按TOP顺序书写，等等\\
\midrule
Radiologic Diagnostic Principles & 影像诊断原则 & e.g., HCC: arterial enhancement, washout, hepatobiliary hypointensity ... \newline 例如HCC：动脉期强化，门脉期或延迟期洗脱，肝胆期低信号等 \\
\midrule
Example Report & 样例报告 & \textbf{Findings:} Liver shows irregular margins and a 4.2×3.3 cm lesion in Segment IVb. Arterial rim enhancement with washout; hypointense on hepatobiliary phase. Enlarged periportal lymph nodes. \newline \textbf{Diagnosis:} 1. Segment IVb lesion suggestive of HCC. 2. Cirrhosis with portal hypertension. 3. Enlarged lymph nodes, metastasis cannot be excluded. \newline \textbf{检查所见：} 肝IVb段见团块状占位，动脉期环形强化，延迟期洗脱，肝胆期低信号，肝门部淋巴结增大。 \newline \textbf{结论：} 1. 肝IVb段占位，考虑HCC可能。2. 肝硬化伴门脉高压。3. 肝门部淋巴结增大，转移不除外。 \\
\bottomrule
\end{tabular}
\label{prompt_summary}
\end{table*}
\end{CJK}

To systematically evaluated the contribution of each component, 11 prompt combinations were developed by selectively integrating different elements from these categories. This design enabled quantitative assessment of how individual and combined prompt features influence the diagnostic reliability, coherence, and consistency of LLM-generated liver MRI reports. Full prompt formulations are provided in the Table~\ref{table1}.

Prompts P0-P6 were designed to test different combinations of instruction-based and example-based components, enabling the identification of the most effective prompt composition strategy. Prompts P5-P11 were further used to examine how the number of examples influences model learning and output performance, thereby clarifying the role of sample augmentation in improving large language model adaptability and diagnostic consistency. We also provided the  information for each prompt's character length in Table~\ref{table1}. 

\begin{table*}[htbp]
\centering
\small
\begin{threeparttable}
\caption{Composition Matrix of the 11 Prompt Configurations.}
\label{table1}
\renewcommand{\arraystretch}{1.2}
\begin{tabular}{
p{1.35cm}  % Prompt ID
p{1.25cm}    % Role Definition
p{1.25cm}  % Core Task Specification
p{1.25cm}  % TOP Diagnostic Taxonomy
p{1.25cm}  % Mandatory Verification Items
p{1.25cm}  % Report Structure Standards
p{1.25cm}  % Imaging Diagnostic Principles
p{1.25cm}  % Number of Examples
p{1.25cm}    % Character Length (Chinese)
}
\toprule
\textbf{Prompt ID} & \textbf{Role Definition} & \textbf{Core Task Specification} &
\textbf{TOP Diagnostic Taxonomy} & \textbf{Mandatory Verification Items} &
\textbf{Report Structure Standards} & \textbf{Imaging Diagnostic Principles} &
\textbf{Number of Examples} & \textbf{Character Length (Chinese)} \\ 
\midrule
Prompt 0 & \checkmark & - & - & - & - & - & 0 & 124\\ 
Prompt 1 & \checkmark & \checkmark & - & - & - & - & 0 & 270\\ 
Prompt 2 & \checkmark & \checkmark & - & - & - & - & 3 & 1411\\ 
Prompt 3 & \checkmark & \checkmark & \checkmark & - & \checkmark & - & 3 & 2036\\ 
Prompt 4 & \checkmark & \checkmark& \checkmark & \checkmark & \checkmark & \checkmark & 3 & 3147\\ 
Prompt 5 & \checkmark & \checkmark & - & - & - & - & 10 & 3946\\ 
Prompt 6 & \checkmark & \checkmark& \checkmark & \checkmark & \checkmark & \checkmark & 10 & 5709\\ 
Prompt 7 & \checkmark & \checkmark& \checkmark & \checkmark & \checkmark & \checkmark & 0 & 2104\\ 
Prompt 8 & \checkmark & \checkmark& \checkmark & \checkmark & \checkmark & \checkmark & 5 & 3953\\ 
Prompt 9 & \checkmark & \checkmark& \checkmark & \checkmark & \checkmark & \checkmark & 15 & 7609\\ 
Prompt 10 & \checkmark & \checkmark& \checkmark & \checkmark & \checkmark & \checkmark & 20 & 9457\\ 
Prompt 11 & \checkmark & \checkmark& \checkmark & \checkmark & \checkmark & \checkmark & 25 & 11025\\ 
\bottomrule
\end{tabular}
\end{threeparttable}
\end{table*}

\subsection{Preparation of Large Language Models}
Four advanced Chinese large language models (LLMs) were evaluated in this study: Kimi-K2-Instruct-0905 (Moonshot AI), Qwen3-235B-A22B-Instruct-2507 (Alibaba Group), DeepSeek-V3 (DeepSeek AI), and ByteDance-Seed-OSS-36B-Instruct (ByteDance). These models represent the leading generation of instruction-tuned Chinese LLMs, each trained on large-scale multimodal or text-based corpora and optimized for complex reasoning and task following \cite{zhang2025igniting,lyu2023faithful}.

To further assess intra-model consistency, the DeepSeek family was additionally tested with its upgraded version DeepSeek-V3.1 and DeepSeek-R1 with Chain-of-Thought \cite{zhao2025insights,chan2025deepseek}. This comparison allowed evaluation of how incremental model updates and explicit reasoning mechanisms affect diagnostic accuracy and report credibility in liver MRI interpretation.

All models were accessed and executed via standardized SiliconFlow APIs to ensure a consistent computational environment and to minimize confounding factors related to hardware or deployment variability. The parameter $enable_thinking$ (if applicable) was uniformly set to $False$ to disable the models' internal "thinking" function, as long conversational contexts were not considered in this study. The parameters $temperature$, $top_p$, and $max_{tokens}$ were set to 0.5, 0.95, and 1024, respectively.

\section{Results}
The results are presented in three parts, corresponding to the experimental workflow. First, DeepSeek-V3 and Kimi-K2 were used to illustrate how different prompt compositions affect model performance. Second, these two models were further applied to evaluate the influence of varying the number of example reports on report quality. Finally, we compared multiple Chinese large language models and validated the DeepSeek family to clarify why DeepSeek-V3 and Kimi-K2 were selected as representative models for subsequent experiments. In addition, we aimed to demonstrate that the designed prompt settings yielded consistent performance patterns across different models, supporting the generalizability of the proposed approach.

\subsection{Effect of Prompt Composition on Model Performance}
In this section, DeepSeek-V3 and Kimi-K2 were evaluated using Prompts P0-P6, as detailed in Table 1. For each prompt configuration, a total of 15,127 liver MRI reports were tested, corresponding to 181,524 chat sessions conducted through the LLM interfaces.

\begin{figure*}[htbp]
  \centerline{\includegraphics[width=0.9\textwidth]{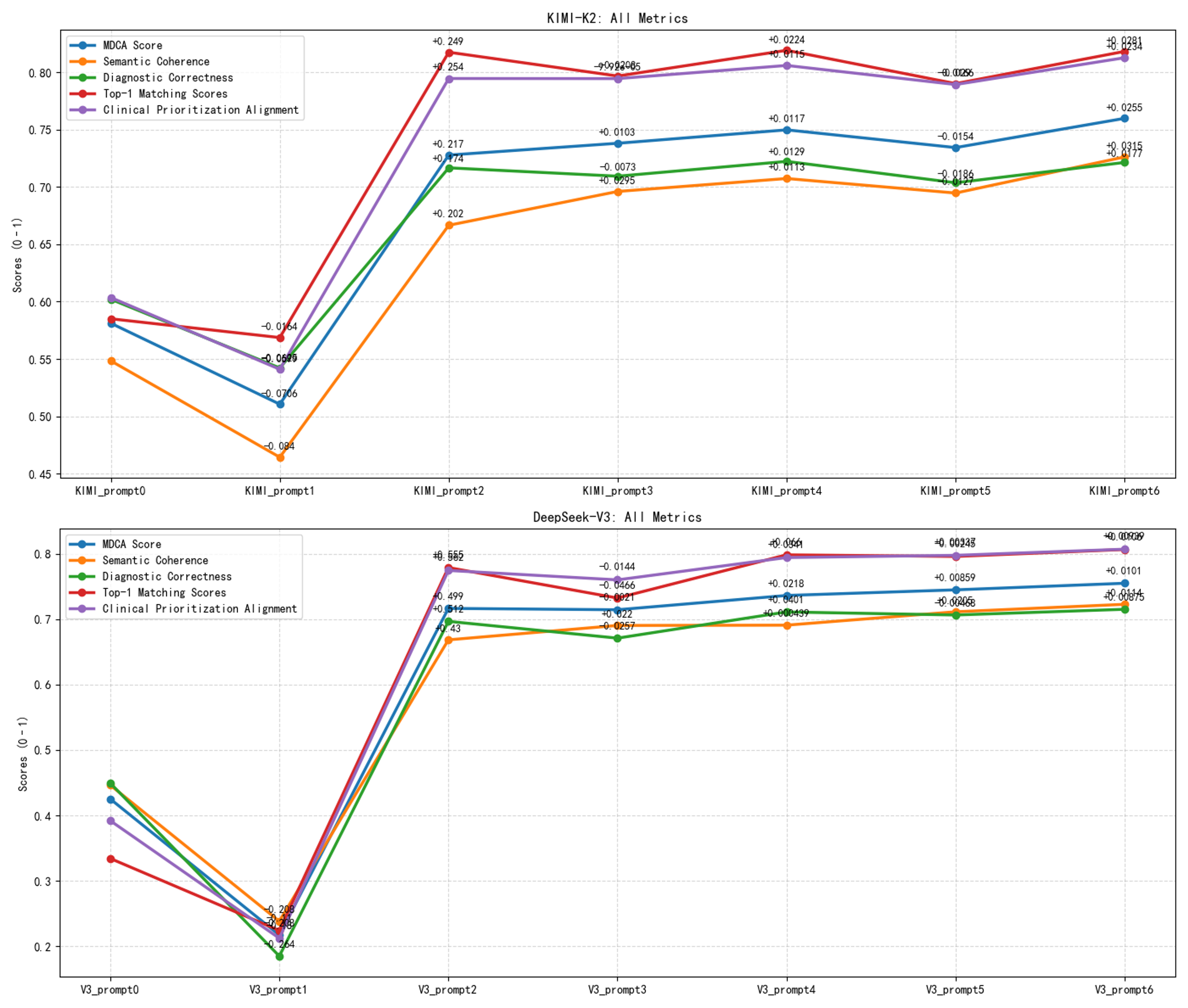}}
  \caption{Experimental results for Prompt Designs with KIMI-K2 and DeepSeek-V3. The figure illustrates the performance of both models across various prompt configurations, highlighting the impact of different prompt components on Semantic Coherence (SC), Diagnostic Correctness (DC), Top-1 matching, Clinical Prioritization Alignment (CPA), and overall MDCA scores.}
  \label{fig2}
  \end{figure*}

As shown in Figure~\ref{fig2} and Table~\ref{table2}, which present the experimental results for prompt designs with Kimi-K2 and DeepSeek-V3, Kimi-K2 consistently outperformed DeepSeek-V3 across all prompt settings. Kimi-K2 with Prompt 6 achieved the highest scores in Semantic Coherence (SC), Clinical Prioritization Alignment (CPA), and overall MDCA, whereas Kimi-K2 with Prompt 4 yielded the best Diagnostic Correctness (DC) and Top-1 matching scores. Performance trends between the two LLMs were highly consistent across all prompt configurations. In addition, higher mean values were accompanied by lower standard deviations, indicating stable model behavior.

Overall, three major observations were identified.

First, prompts that included role definition but lacked explicit diagnostic guidance performed poorly, and those containing only role definition and core task specification without structured support achieved even lower scores (Prompts 0 and 1).

Second, compared with instruction-only prompts, the inclusion of example-based guidance substantially improved overall performance (Prompts 1, 2, 5, and 4, 6), although this required greater token consumption and communication overhead.

Finally, the integration of key instruction-based components—specifically the tiered diagnostic taxonomy, mandatory verification items, report structure standards, and imaging diagnostic principles—further enhanced SC, DC, and Top-1 matching scores, leading to more accurate and diagnostically reliable outputs (Prompts 3, 4, and 5, 6).

  \begin{table*}[htbp]
    \centering
    \small
    \begin{threeparttable}
    \caption{Evaluation Metrics for KIMI-K2 and DeepSeek-V3 with Different Prompt Designs.}
    \label{table2}
    \renewcommand{\arraystretch}{1.2}
    \begin{tabular}{
    p{1.35 cm}  % Prompt ID
    p{1.35cm}    % Role Definition
    p{1.35cm}  % Core Task Specification
    p{1.35cm}  % TOP Diagnostic Taxonomy
    p{1.35cm}  % Mandatory Verification Items
    p{1.35cm}  % Report Structure Standards
    p{1.35cm}  % Imaging Diagnostic Principles
    p{1.35cm}  % Number of Examples
    }
    \toprule
    \textbf{KIMI Mean(STD)} & \textbf{Prompt 0} & \textbf{Prompt 1} &
    \textbf{Prompt 2} & \textbf{Prompt 3} &
    \textbf{Prompt 4} & \textbf{Prompt 5}  & \textbf{Prompt 6} \\ 
    \midrule
    SC & 0.5483 (0.2651) & 0.4643 (0.251) & 0.6666 (0.1966) &0.6961 (0.1962) &0.7073 (0.1925) &0.6947 (0.1967) &0.7262 (0.191) \\ 
    DC & 0.6021 (0.3352) & 0.5422 (0.3708) & 0.7167 (0.2752) & 0.7094 (0.2799) & 0.7223 (0.2725) & 0.7036 (0.2839) & 0.7214 (0.2742) \\ 
    Top-1 & 0.5851 (0.4927) & 0.5687 (0.4953) & 0.8175 (0.3863) & 0.7967 (0.4025) & 0.8191 (0.3849) & 0.7901 (0.4073) & 0.8182 (0.3857) \\ 
    CPA & 0.6035 (0.3995) & 0.541 (0.41) & 0.7945 (0.2982) & 0.7944 (0.3028) & 0.8059 (0.2892) & 0.7893 (0.3019) & 0.8127 (0.2867) \\ 
MDCA Score & 0.5812 (0.2834)& 0.5106 (0.2878)& 0.7278 (0.1907)& 0.7381 (0.1954)& 0.7498 (0.1865)& 0.7343 (0.1949)& 0.7598 (0.1869) \\
    \midrule
    \textbf{DeepSeek Mean(STD)} & \textbf{Prompt 0} & \textbf{Prompt 1} &
    \textbf{Prompt 2} & \textbf{Prompt 3} &
    \textbf{Prompt 4} & \textbf{Prompt 5}  & \textbf{Prompt 6} \\ 
    \midrule
    SC & 0.4462 (0.2436)& 0.2381 (0.3157)& 0.6683 (0.1952)& 0.6903 (0.1952)& 0.6908 (0.1948)& 0.7112 (0.1947)& 0.7227 (0.1914) \\
    DC &0.4492 (0.3596)& 0.1848 (0.3065)& 0.6966 (0.2879)& 0.6709 (0.2995)& 0.711 (0.282)& 0.7063 (0.2844)& 0.7151 (0.2806) \\
    Top-1 & 0.334 (0.4717)& 0.2239 (0.4169)& 0.7789 (0.415)& 0.7323 (0.4428)& 0.7983 (0.4013)& 0.7959 (0.4031)& 0.8065 (0.3951) \\
    CPA & 0.3918 (0.3804)& 0.2121 (0.3612)& 0.7744 (0.3128)& 0.76 (0.3232)& 0.7942 (0.3009)& 0.7975 (0.2976)& 0.8069 (0.2927) \\
MDCA Score & 0.4250 (0.2667)& 0.2170 (0.3066)& 0.7164 (0.1973)& 0.7143 (0.2051)& 0.7362 (0.1922)& 0.7448 (0.1928)& 0.7548 (0.1894) \\
    \bottomrule
    \end{tabular}
    \end{threeparttable}
    \end{table*}

\subsection{Effect of Example Number on Model Performance}
\begin{figure*}[htbp]
  \centerline{\includegraphics[width=0.9\textwidth]{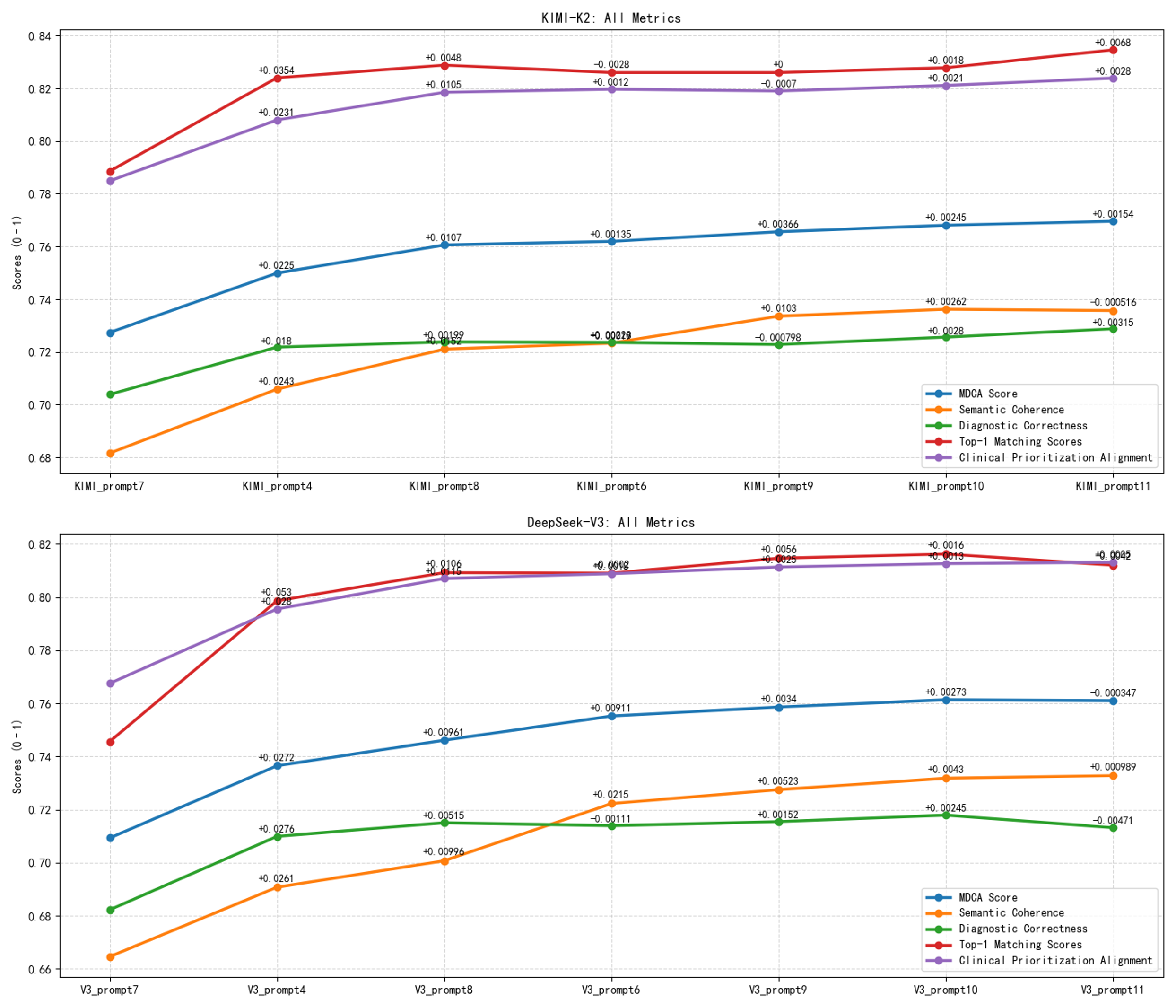}}
  \caption{Experimental Results with Different Example Numbers. The figure illustrates how varying the number of example reports in the prompts affects the performance of KIMI-K2 and DeepSeek-V3 across multiple evaluation metrics, including Semantic Coherence (SC), Diagnostic Correctness (DC), Top-1 matching, Clinical Prioritization Alignment (CPA), and overall MDCA scores.}
  \label{fig3}
  \end{figure*}

In this section, DeepSeek-V3 and Kimi-K2 were evaluated using Prompts 4 and 6-11. When sorted by the number of examples, these correspond to Prompts 7, 4, 8, 6, 9, 10, and 11, which contained 0, 3, 5, 10, 15, and 20 examples, respectively. For each prompt configuration, a total of 5,000 liver MRI reports were tested, corresponding to 70,000 chat sessions conducted through the LLM interfaces. The experimental results are summarized in Figure~\ref{fig3} and Table~\ref{table3}. 

Both Kimi-K2 and DeepSeek-V3 exhibited consistent improvement across all evaluation dimensions as the number of examples increased. Performance gains were most notable in Semantic Coherence (SC) and Top-1 matching accuracy, indicating that additional examples enhanced the models' ability to generate fluent, contextually coherent, and diagnostically aligned reports.

Model performance increased steadily up to approximately 20 examples, beyond which further expansion yielded marginal or slightly negative effects. The composite score for DeepSeek-V3 peaked at 0.7613 (Prompt 10), while Kimi-K2 reached 0.7696 (Prompt 11), after which performance plateaued. Overall, the improvements beyond 20 examples were limited. These findings suggest that while moderate example augmentation—around 10 to 20 examples—substantially improves language fluency and diagnostic completeness, excessive examples may introduce contextual noise and redundancy that hinder performance.

Notably, both models followed highly similar trends across all metrics, confirming that the effect of example number is consistent and model-agnostic. This consistency suggests that the observed performance gains stem from the underlying optimization of contextual understanding rather than model-specific characteristics, thereby supporting the generalizability of this prompt-scaling strategy across Chinese LLMs.

Finally, when instruction-based components are limited, we believe expanding the number of examples provides an effective alternative for improving performance, albeit with higher token and communication costs. In practical deployment, we recommend using a comprehensive instruction-based prompt design combined with approximately 10-15 representative examples, which offers an optimal balance between performance, efficiency, and feasibility for most clinical institutions.

  \begin{table*}[htbp]
    \centering
    \small
    \begin{threeparttable}
    \caption{Evaluation Metrics for KIMI-K2 and DeepSeek-V3 with Different Example Numbers.}
    \label{table3}
    \renewcommand{\arraystretch}{1.2}
    \begin{tabular}{
    p{1.35 cm}  % Prompt ID
    p{1.35cm}    % Role Definition
    p{1.35cm}  % Core Task Specification
    p{1.35cm}  % TOP Diagnostic Taxonomy
    p{1.35cm}  % Mandatory Verification Items
    p{1.35cm}  % Report Structure Standards
    p{1.45cm}  % Imaging Diagnostic Principles
    p{1.45cm}  % Number of Examples
    }
    \toprule
    \textbf{KIMI Mean(STD)} & \textbf{Prompt 7} & \textbf{Prompt 4} &
    \textbf{Prompt 8} & \textbf{Prompt 6} &
    \textbf{Prompt 9} & \textbf{Prompt 10}  & \textbf{Prompt 11} \\ 
    \midrule
    SC & 0.6816 (0.1934)& 0.7058 (0.1925)& 0.721 (0.1906)& 0.7233 (0.1922)& 0.7335 (0.1898)& 0.7362 (0.189)& 0.7357 (0.1887) \\
    DC &0.7038 (0.2826)& 0.7218 (0.2701)& 0.7238 (0.2704)& 0.7236 (0.2697)& 0.7228 (0.2716)& 0.7256 (0.2692)& 0.7287 (0.2682) \\
    Top-1 & 0.7886 (0.4084)& 0.824 (0.3809)& 0.8288 (0.3768)& 0.826 (0.3792)& 0.826 (0.3792)& 0.8278 (0.3776)& 0.8346 (0.3716) \\
    CPA & 0.7849 (0.3084)& 0.808 (0.2891)& 0.8185 (0.2822)& 0.8197 (0.2842)& 0.819 (0.2824)& 0.8211 (0.2808)& 0.8239 (0.2796) \\ 
MDCA Score &0.7273 (0.1981) & 0.7500 (0.1856) & 0.7605 (0.1834) & 0.7619 (0.1852) & 0.7656 (0.1839) & 0.7680 (0.1826) &0.7696 (0.1812) \\
    \midrule
    \textbf{DeepSeek Mean(STD)} & \textbf{Prompt 7} & \textbf{Prompt 4} &
    \textbf{Prompt 8} & \textbf{Prompt 6} &
    \textbf{Prompt 9} & \textbf{Prompt 10}  & \textbf{Prompt 11} \\ 
    \midrule
    SC & 0.6646 (0.194) & 0.6907 (0.1943) & 0.7007 (0.1927) & 0.7222 (0.1895) & 0.7275 (0.1923) & 0.7318 (0.1918) & 0.7328 (0.1935) \\
    DC &0.6822 (0.2932) & 0.7099 (0.2805) & 0.715 (0.2789) & 0.7139 (0.2796) & 0.7154 (0.2775) & 0.7179 (0.2765) & 0.7132 (0.279) \\
    Top-1 & 0.7456 (0.4356) & 0.7986 (0.4011) & 0.8092 (0.393) & 0.809 (0.3931) & 0.8146 (0.3887) & 0.8162 (0.3874) & 0.812 (0.3908) \\
    CPA & 0.7675 (0.3189) & 0.7955 (0.2982) & 0.807 (0.2929) & 0.8088 (0.2916) & 0.8113 (0.2884) & 0.8126 (0.2871) & 0.8131 (0.2885) \\
MDCA Score & 0.7093 (0.2041) & 0.7365 (0.1902) & 0.7461 (0.1886) & 0.7552 (0.1881) & 0.7586 (0.1874) & 0.7613 (0.1865) & 0.7610 (0.1889) \\
    \bottomrule
    \end{tabular}
    \end{threeparttable}
    \end{table*}

\subsection{Comparison of Chinese LLMs and Validation within the DeepSeek Family}
\begin{figure*}[htbp]
  \centerline{\includegraphics[width=0.9\textwidth]{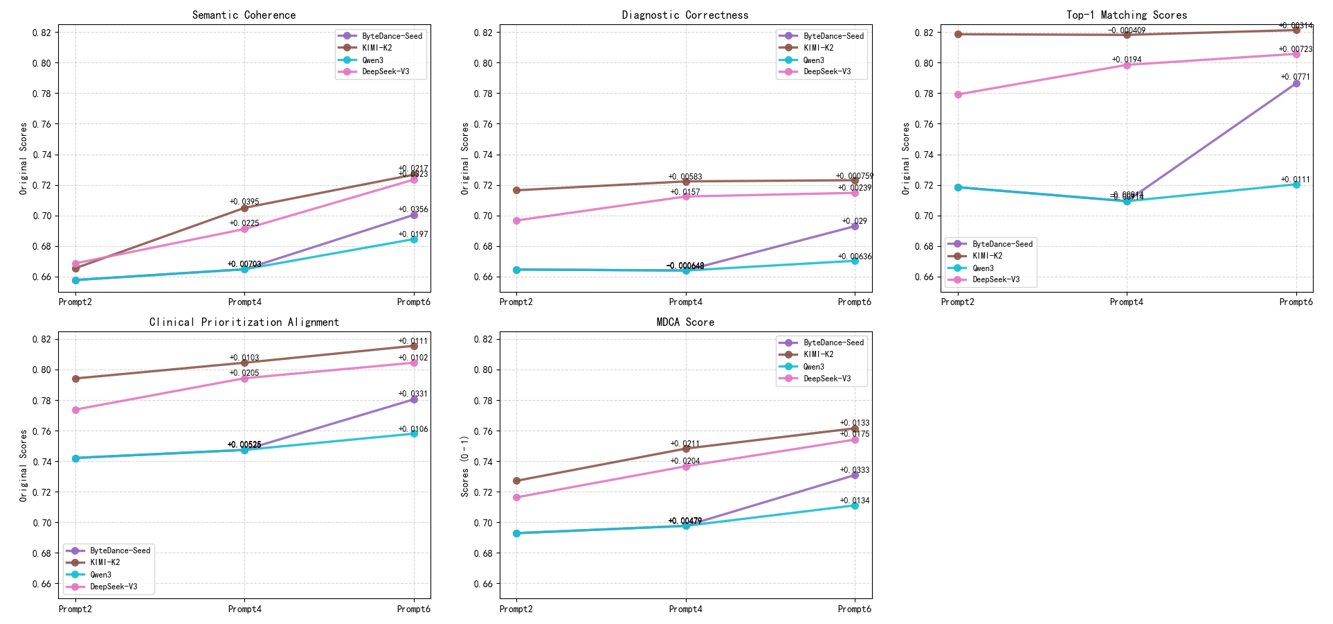}}
  \caption{Experimental Results with Four LLMs. The figure illustrates the performance of KIMI-K2, DeepSeek-V3, ByteDance-Seed, and Qwen3-235B across various prompt configurations.}
  \label{fig4}
  \end{figure*}

In this section, four major Chinese large language models—Kimi-K2-Instruct-0905, DeepSeek-V3, ByteDance-Seed-OSS-36B-Instruct, and Qwen3-235B-A22B-Instruct-2507—were evaluated under prompt configurations of varying complexity (Prompts 2, 4, and 6). For each model-prompt combination, 7,500 liver MRI reports were tested, resulting in approximately 90,000 chat sessions conducted through the SiliconFlow platform. Experimental results are summarized in Figure~\ref{fig4}.

All models demonstrated consistent performance gains with increasing prompt sophistication, highlighting the crucial role of structured instruction and contextual guidance. Kimi-K2 achieved the best overall performance, with composite scores improving from 0.7271 (Prompt 2) to 0.7615 (Prompt 6). DeepSeek-V3 followed closely, reaching 0.7541 under the same condition. In contrast, ByteDance-Seed and Qwen3-235B showed limited improvements, particularly in Semantic Coherence (SC) and Clinical Prioritization Alignment (CPA), indicating weaker adaptability to complex prompt instructions.

\begin{figure*}[htbp]
  \centerline{\includegraphics[width=0.9\textwidth]{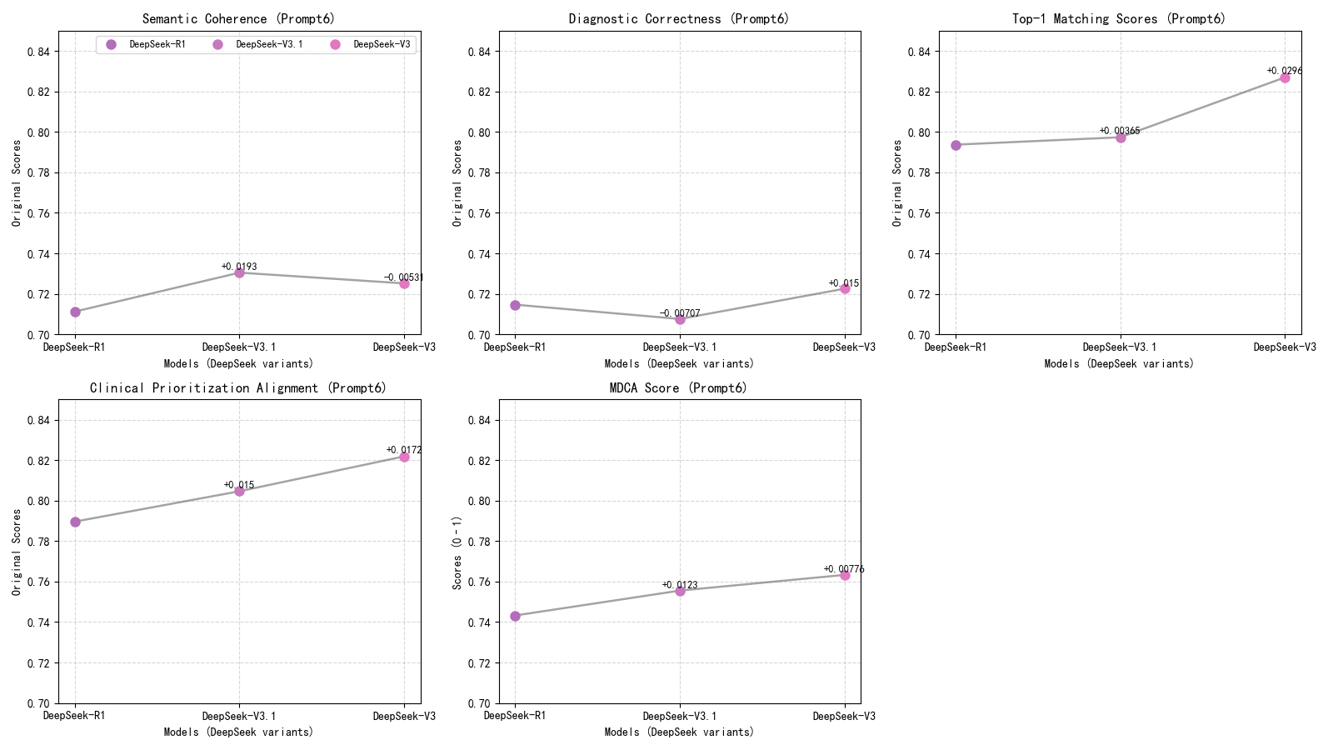}}
  \caption{Experimental Results with DeepSeek Family. The figure illustrates the performance of DeepSeek-V3, DeepSeek-V3.1, and DeepSeek-R1 across various prompt configurations.}
  \label{fig5}
  \end{figure*}

Further validation within the DeepSeek family was conducted using 2,500 liver MRI reports per model-prompt configuration, enabling detailed comparison among DeepSeek-V3, DeepSeek-V3.1, and DeepSeek-R1. As shown in Figure~\ref{fig5}, DeepSeek-V3 achieved the highest composite score (0.7633), outperforming V3.1 (0.7555) and R1 (0.7433). While V3.1 exhibited marginally higher Semantic Coherence (0.730 vs. 0.725), V3 demonstrated superior Diagnostic Correctness (DC), Top-1 matching accuracy, and CPA, producing more diagnostically reliable reports. The reasoning-augmented R1 achieved slightly higher semantic correctness (0.715) but at the cost of reduced consistency and fluency. These findings indicate that explicit Chain-of-Thought reasoning does not necessarily enhance clinical reliability, and that optimized instruction design remains more impactful for structured diagnostic generation.

Overall, both Kimi-K2 and DeepSeek-V3 emerged as the most reliable and balanced performers, consistently benefiting from prompt and sample optimization. The performance patterns across all models—characterized by steady improvement with increased instruction richness—demonstrate the model-agnostic generalizability of the proposed prompt customization and credibility evaluation frameworks. Consequently, these two models were selected as representative systems for subsequent experiments, reflecting their strong baseline capability, responsiveness to structured prompting, and robustness across diverse evaluation dimensions.

\subsection{Cross-Center Validation on External Liver MRI Cohort}
To evaluate whether the proposed prompt optimization strategy generalizes across 
institutions, an independent external cohort of 1,009 liver MRI reports was tested 
using Kimi-K2 and DeepSeek-V3. Beyond directly transferring the original prompts, 
we further examined a practical deployment scenario: when moving to a new center, 
the terminology and tiered diagnostic taxonomy (TOP system) were modestly revised 
to align with local reporting conventions, and all example reports were replaced 
with those sourced from the target center.

Five prompt variants were compared:
\begin{itemize}
  \setlength{\itemsep}{0pt}
  \item \textbf{Prompt~7} and \textbf{Prompt~11}: the original instruction-based 
    prompt (zero examples) and the full-configuration prompt (25 primary-center examples), 
    respectively, as defined in Table~\ref{table1};
  \item \textbf{Prompt~11 V2}: Prompt~11 with revised terminology and diagnostic taxonomy 
    adapted to the external center;
  \item \textbf{Prompt~7 V2}: Prompt~11 V2 with all example reports removed, i.e., 
    adapted instructions only, zero examples;
  \item \textbf{Prompt~11 V3}: Prompt~7 V2 with all example reports replaced by 
    25 external-center samples.
\end{itemize}

The results are summarized in Table~\ref{table4}. Three key findings emerged.

First, directly transplanting the original Prompt~7 and Prompt~11 to the external 
center yielded MDCA scores of 0.615--0.673 (Kimi-K2) and 0.651--0.691 (DeepSeek-V3), 
confirming that the proposed framework retained reasonable performance without any 
center-specific adaptation. However, these scores were notably lower than those 
observed on the primary center (\textit{cf.} Table~\ref{table2}), reflecting 
inter-institutional differences in reporting style and diagnostic phrasing.

Second, adapting terminology and replacing examples with local samples 
(Prompt~11 V3) produced the highest scores across both models: MDCA reached 
0.774 (Kimi-K2) and 0.768 (DeepSeek-V3). Importantly, even terminology adaptation 
alone without examples (Prompt~7 V2) yielded competitive performance (Kimi-K2: 0.703; 
DeepSeek-V3: 0.692), approaching or surpassing the original Prompt~11 with 
primary-center examples.

Third, DeepSeek-V3 exhibited a higher zero-shot baseline than Kimi-K2 under the 
original Prompt~7 (MDCA: 0.651 \textit{vs.} 0.615), whereas Kimi-K2 responded more 
substantially to example augmentation and terminology refinement, ultimately 
achieving the highest composite score.

These results demonstrate that institution-specific prompt adaptation—combining 
modest terminology revision with locally sourced example reports—enables high-quality 
LLM-generated reports in a new center at relatively low adaptation cost, avoiding 
the need to redesign the entire prompt from scratch.

\begin{table*}[htbp]
  \centering
  \small
  \begin{threeparttable}
  \caption{Cross-Center Validation Results on the External Liver MRI Cohort (n = 1,009).}
  \label{table4}
  \renewcommand{\arraystretch}{1.2}
  \begin{tabular}{
  p{2.5cm}  % Label
  p{1.45cm}  % Prompt 7
  p{1.45cm}  % Prompt 11
  p{1.85cm}  % Prompt 11 V2
  p{1.85cm}  % Prompt 7 V2
  p{2.2cm}  % Prompt 11 V3
  }
  \toprule
  Kimi-K2 (mean\_std)& \textbf{Prompt~7} & \textbf{Prompt~11} & \textbf{Prompt~11 V2} & \textbf{Prompt~7 V2} & \textbf{Prompt~11 V3} \\
  \midrule
  SC      & 0.7057 (0.1769) & 0.7634 (0.1547) & 0.7894 (0.1415) & 0.6944 (0.1816) & 0.7883 (0.1448) \\
  DC      & 0.4793 (0.3002) & 0.5210 (0.2940) & 0.5788 (0.2775) & 0.6499 (0.2855) & 0.6926 (0.2703) \\
  Top-1   & 0.6868 (0.4640) & 0.7889 (0.4083) & 0.8474 (0.3598) & 0.7750 (0.4178) & 0.8583 (0.3489) \\
  CPA     & 0.5927 (0.3509) & 0.6586 (0.3275) & 0.7289 (0.2948) & 0.7374 (0.3156) & 0.7993 (0.2784) \\
  MDCA    & 0.6152 (0.2093) & 0.6730 (0.1950) & 0.7231 (0.1724) & 0.7027 (0.1963) & 0.7736 (0.1722) \\
  \midrule
  DeepSeek-V3 (mean\_std)& \textbf{Prompt~7} & \textbf{Prompt~11} & \textbf{Prompt~11 V2} & \textbf{Prompt~7 V2} & \textbf{Prompt~11 V3} \\
  \midrule
  SC      & 0.6713 (0.1952) & 0.7682 (0.1511) & 0.8011 (0.1460) & 0.6713 (0.1952) & 0.8011 (0.1460) \\
  DC      & 0.5535 (0.2944) & 0.5466 (0.2847) & 0.5576 (0.2917) & 0.6361 (0.2896) & 0.6639 (0.2838) \\
  Top-1   & 0.7820 (0.4131) & 0.8246 (0.3805) & 0.8216 (0.3830) & 0.7849 (0.4111) & 0.8266 (0.3788) \\
  CPA     & 0.6804 (0.3276) & 0.6868 (0.3142) & 0.7071 (0.3105) & 0.7398 (0.3061) & 0.7874 (0.2829) \\
  MDCA    & 0.6514 (0.2034) & 0.6913 (0.1869) & 0.7148 (0.1881) & 0.6917 (0.1983) & 0.7682 (0.1792) \\
  \bottomrule
  \end{tabular}
  \begin{tablenotes}
  \item Prompt~11 V2 revises the tiered diagnostic taxonomy and 
  terminology for the external center; Prompt~7 V2 removes all examples from Prompt~11 V2; 
  Prompt~11 V3 replaces all examples with external-center samples. Columns~4--5 use the 
  updated scoring system with synonym expansion.
  \end{tablenotes}
  \end{threeparttable}
  \end{table*}

  \subsection{Cross-Disease Validation on Lung Cancer}
  To assess cross-disease generalizability, the same six-component prompt architecture 
  (Table~\ref{table6}) was applied to pulmonary thin-section CT reporting 
  at the primary center, using 1692 lung cancer reports processed with Kimi-K2 under Prompt~0, Prompt~7 and Prompt~10.

\begin{table}[htbp]
  \centering
  \small
  \begin{threeparttable}
  \caption{Cross-Disease Validation Results on Lung Cancer (n = 1692).}
  \label{table6}
  \renewcommand{\arraystretch}{1.2}
  \begin{tabular}{
    p{1.2 cm}
    p{1.4cm}
    p{1.4cm}
    p{1.4cm}
    }
    \toprule
    Kimi-K2 & \textbf{Prompt~0} & \textbf{Prompt~7} & \textbf{Prompt~10} \\
    \midrule
    SC   & 0.767 (0.1861)    & 0.7506 (0.1279)    & 0.7987 (0.1309) \\
    DC   & 0.334 (0.2522)    & 0.5304 (0.2705)    & 0.6237 (0.2541) \\
    Top-1& 0.3182 (0.4659)   & 0.6339 (0.4819)    & 0.7753 (0.4175) \\
    CPA  & 0.2593 (0.3149)   & 0.5455 (0.3295)    & 0.6668 (0.3048) \\
    MDCA & 0.4773 (0.1701) & 0.6245 (0.1788) & 0.7109 (0.1676) \\
    \bottomrule
  \end{tabular}
  \end{threeparttable}
\end{table}

  Although structurally identical to the liver MRI prompt, the pulmonary version was 
  inherently more complex in several respects. Anatomically, it spanned seven subsystems 
  (lung parenchyma, airways, pleura, mediastinum, great vessels, chest wall, and esophagus) 
  rather than a single organ. It required simultaneous adherence to five external 
  standardization systems (Lung-RADS, Boyden segmental nomenclature, IASLC 9th-edition 
  lymph node station map, three-segment esophageal classification, and rib/vertebral 
  segmental localization), compared with essentially one (Couinaud anatomy) for liver MRI. 
  Furthermore, the pulmonary prompt incorporated a critical-value alerting mechanism 
  for life-threatening conditions (e.g., aortic dissection, tension pneumothorax), 
  a feature absent from the liver protocol. The broader disease spectrum and greater 
  proportion of diagnostically indeterminate lesions also necessitated stricter 
  anti-hallucination safeguards and produced substantially higher prompt information 
  density than its hepatic counterpart.
  
  Despite these differences, the six-component framework transferred directly without 
  architectural modification, yielding consistent performance improvements on lung 
  cancer reports. These results confirm that the proposed 
  prompt design and MDCA evaluation frameworks generalize effectively across distinct 
  disease domains, with the principal adaptation cost being content expansion of 
  existing components rather than structural redesign.

\subsection{Consistency Check between MDCA Framework and Radiologist Judgments}
To evaluate the clinical alignment of the MDCA framework, we conducted a consistency check between MDCA scores and radiologist assessments. Four representative prompts (Prompt 0, 2, 4, and 6) from Kimi-K2 were selected for comparison. Two board-certified radiologists independently reviewed 100 reports per prompt, totaling 800 reports, comparing each LLM-generated report against the corresponding ground-truth report.

Each report was rated using a three-tier scale:

Grade A (1 point): Fully acceptable, with clinically sound content and no major errors;
Grade B (0.5 points): Acceptable with minor, tolerable issues (e.g., slightly ambiguous phrasing or minor inconsistencies);
Grade C (0 points): Unacceptable due to incoherent language or critical factual inaccuracies
Each report received two individual ratings, and the combined score served as the final consistency judgment. Scores of 2.0 were considered excellent, 1.0-1.5 acceptable, and $<1$ unacceptable.

\begin{figure}[htbp]
  \centerline{\includegraphics[width=0.5\textwidth]{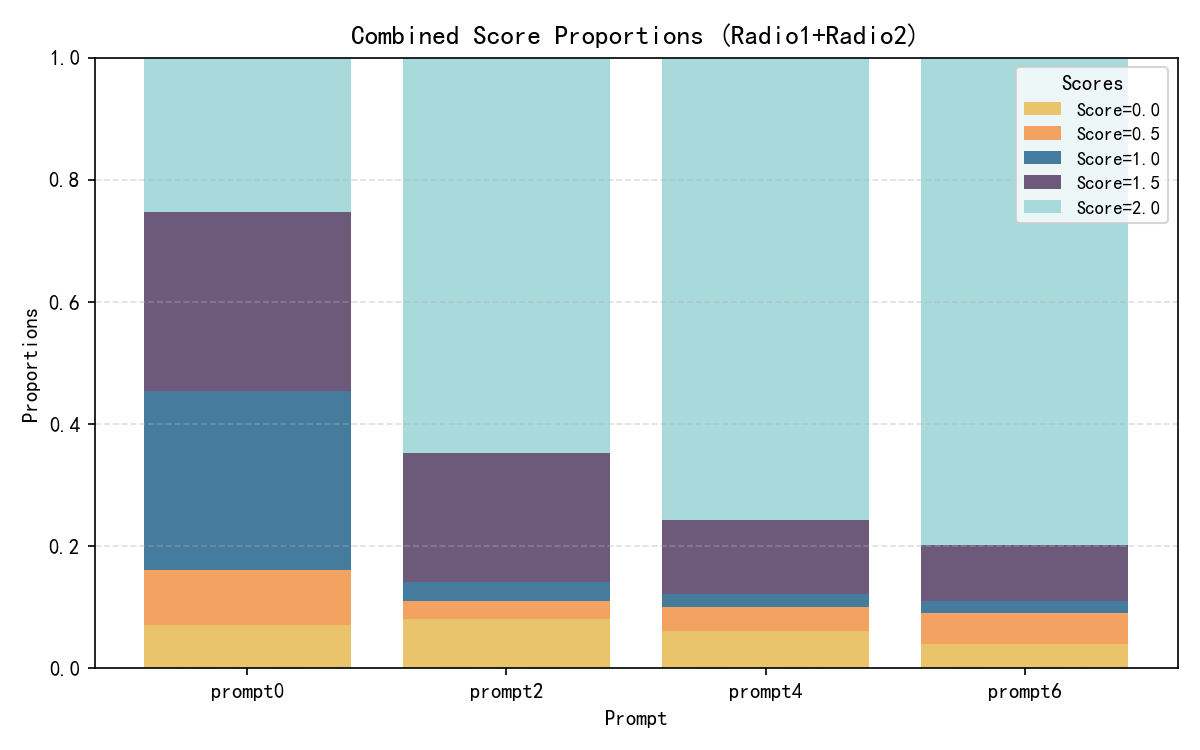}}
  \caption{Combined Scores and Proportions of Radiologist Ratings for KIMI-K2 under Different Prompts. The figure illustrates the distribution of radiologist ratings (Grade A, B, C) for KIMI-K2-generated reports across four prompt configurations (Prompt 0, 2, 4, and 6), highlighting the correlation between prompt quality and clinical acceptability.}
  \label{fig6}
  \end{figure}

As illustrated in Figure~\ref{fig6}, the proportion of reports rated as excellent (score = 2.0) increased substantially with improved prompt quality, from Prompt 0 to Prompt 6. In contrast, lower-quality prompts yielded a higher proportion of partially or wholly unacceptable outputs. These trends closely aligned with the MDCA scores, supporting the clinical validity and interpretability of the proposed framework.

Manual consistency checks, however, are labor-intensive and not scalable at the population level, further underscoring the need for automated, reproducible frameworks such as MDCA.

  \section{Discussion}
  \subsection{Prompt Design Principles and Mechanistic Insights}
  Our findings highlight that prompt construction in large language model (LLM)-based radiology reporting is not merely additive but synergistic. Instruction-based components—including role definition, core task specification, tiered diagnostic taxonomy (TOP system), mandatory verification items, report structure standards, and imaging diagnostic principles—substantially improved Diagnostic Correctness (DC) and Clinical Prioritization Alignment (CPA) by enforcing structured reasoning and domain-specific consistency. In contrast, the inclusion of example-based guidance primarily enhanced Semantic Coherence (SC), promoting fluency and logical continuity in report narratives. However, excessive examples ($>20$) introduced contextual redundancy and token interference, resulting in performance saturation or mild decline. These results suggest that an optimal balance—combining comprehensive instruction design with approximately 10--15 representative examples—achieves the best trade-off between diagnostic accuracy, fluency, and computational efficiency. Notably, cross-center validation further revealed that replacing primary-center examples with locally sourced samples from the target institution yielded additional performance gains, underscoring the value of institution-specific example curation in multi-center deployment.
  
  \subsection{Model Performance and Selection Considerations}
  Among the evaluated LLMs, Kimi-K2 and DeepSeek-V3 consistently outperformed other Chinese LLMs, exhibiting strong baseline reliability and responsiveness to prompt optimization. Within the DeepSeek family, V3 achieved the most balanced and diagnostically consistent performance, surpassing both V3.1 and the reasoning-augmented R1 variant. While V3.1 showed slightly higher semantic coherence, and R1 demonstrated marginal gains in correctness, neither delivered stable improvements across diagnostic metrics. These findings indicate that the integration of explicit reasoning mechanisms does not necessarily translate to higher clinical reliability, whereas structured prompt optimization has a more direct and reproducible effect. Cross-center experiments further showed that DeepSeek-V3 maintained a stronger performance, while Kimi-K2 responded more substantially to institution-specific adaptation, suggesting complementary strengths for different deployment scenarios.
  
  Importantly, the consistent response patterns observed across all models—extending from the primary-center liver cohort to the external-center liver cohort and the cross-disease lung validation—demonstrate the generalizability of the proposed prompt design protocols. This cross-model, cross-center, and cross-disease stability suggests that scientifically designed prompts exert a greater influence on report quality than model selection itself. In other words, the reliability and trustworthiness of LLM-generated radiology reports depend more on how models are instructed than on which models are used—underscoring the central role of prompt engineering in clinical LLM deployment.
  
  \subsection{Practical Applications and Clinical Implications}
  Beyond technical performance, the proposed frameworks hold strong potential for real-world applications in radiology workflows. The Multi-Dimensional Credibility Assessment (MDCA) system provides an interpretable and quantitative basis for quality control, allowing automated benchmarking of report consistency, diagnostic accuracy, and adherence to institutional standards. Such a framework could be integrated into hospital reporting systems to assist in automated quality assurance and longitudinal monitoring of reporting performance.
  
  The MDCA framework offers several distinct advantages over existing single-metric or LLM-based self-evaluation approaches. First, it evaluates model outputs across three complementary dimensions—Semantic Coherence (SC), Diagnostic Correctness (DC), and Clinical Prioritization Alignment (CPA)—capturing both linguistic and clinical reliability within a unified structure; Second, the framework is model-agnostic, enabling fair comparison among different LLMs without dependence on internal architectures or training data; Third, each dimension is interpretable and directly maps to radiological reasoning processes, allowing human experts to trace and validate model errors. Together, these features make MDCA a practical, transparent, and reproducible tool for continuous model evaluation and regulatory oversight. The successful transfer of the MDCA framework from liver MRI to pulmonary CT further confirms its disease-agnostic design, requiring only content-level adaptation of the scoring rubrics rather than architectural revision.
  
  From an educational perspective, the prompt optimization and credibility evaluation frameworks can also support radiology trainee education. LLM-generated reports, when aligned with expert-verified prompts, can serve as high-quality exemplars to improve trainees' diagnostic reasoning, linguistic precision, and report standardization. This dual application—quality control and education—demonstrates the broader clinical and pedagogical value of trustworthy LLM deployment in medical imaging.
  
  \section{Limitations}
  This study has several limitations.
  
  First, it focused exclusively on text-to-text large language models (LLMs) rather than multimodal systems capable of directly processing imaging data. While multimodal models hold promise for integrating visual and textual reasoning, they currently face deployment barriers related to high GPU requirements, complex DICOM handling, and limited clinical infrastructure. In contrast, text-based frameworks such as DeepSeek-V3 and Kimi-K2 provide a more practical and scalable alternative, particularly for resource-constrained healthcare institutions.
  
  Second, the evaluation involved only radiologists, without incorporating feedback from other clinical specialists such as surgeons or oncologists. Although this focus ensured domain consistency, it may have limited the comprehensiveness of the credibility assessment (MDCA) in capturing broader multidisciplinary perspectives. Future studies could include multi-specialty panels to refine and expand the interpretive dimensions of the MDCA framework.
  
  Third, although cross-center validation was conducted on an external liver MRI cohort, and cross-disease testing was performed on lung cancer reports, both the primary-center lung experiments and the external-center liver experiments were each limited to a single institution. Model behaviors and prompt effectiveness could still vary across additional centers with different reporting styles and linguistic norms. Larger multicenter and multi-disease longitudinal studies are warranted to further validate the reproducibility and scalability of the proposed frameworks in diverse real-world settings.
  
  \section{Conclusions}
  This study introduces a Multi-Dimensional Credibility Assessment (MDCA) framework that enables objective, transparent, and reproducible evaluation of LLM-generated liver MRI reports. Using this framework, we systematically investigated how scientific prompt design and model selection influence report quality. Cross-center validation on an external liver MRI cohort and cross-disease testing on lung cancer reports confirmed the generalizability of both the prompt design protocol and the MDCA evaluation framework. The findings indicate that structured instruction with approximately 10--15 representative examples—preferably sourced locally—achieves the most balanced and reliable performance across institutions and disease contexts. Among the evaluated models, Kimi-K2 and DeepSeek-V3 consistently demonstrated superior overall credibility. These results underscore that both rational prompt engineering and credible model evaluation are essential for integrating trustworthy large language models into radiology workflows for quality assurance and professional training.

\bibliographystyle{IEEEtran}
\bibliography{ijcai20}

@article{chen2025fuzzy,
  title={Fuzzy Reasoning Chain: An Innovative Reasoning Framework from Fuzziness to Clarity},
  author={Chen, Ping and Liu, Xiang and Liu, Zhaoxiang and Chen, Zezhou and Zhang, Xingpeng and Hu, Huan and Wang, Zipeng and Wang, Kai and Shi, Shuming and Lian, Shiguo},
  journal={arXiv preprint arXiv:2509.22054},
  year={2025}
}

@article{llovet2022immunotherapies,
  title={Immunotherapies for hepatocellular carcinoma},
  author={Llovet, Josep M and Castet, Florian and Heikenwalder, Mathias and Maini, Mala K and Mazzaferro, Vincenzo and Pinato, David J and Pikarsky, Eli and Zhu, Andrew X and Finn, Richard S},
  journal={Nature reviews Clinical oncology},
  volume={19},
  number={3},
  pages={151--172},
  year={2022},
  publisher={Nature Publishing Group UK London}
}

@article{bruix2005management,
  title={Management of hepatocellular carcinoma},
  author={Bruix, Jordi and Sherman, Morris},
  journal={Hepatology},
  volume={42},
  number={5},
  pages={1208--1236},
  year={2005},
  publisher={Wiley Online Library}
}

@article{VOGEL20221345,
title = {Hepatocellular carcinoma},
journal = {The Lancet},
volume = {400},
number = {10360},
pages = {1345-1362},
year = {2022},
issn = {0140-6736},
doi = {https://doi.org/10.1016/S0140-6736(22)01200-4},
url = {https://www.sciencedirect.com/science/article/pii/S0140673622012004},
author = {Arndt Vogel and Tim Meyer and Gonzalo Sapisochin and Riad Salem and Anna Saborowski},
}

@article{aoki2021higher,
  title={Higher enhancement intrahepatic nodules on the hepatobiliary phase of Gd-EOB-DTPA-enhanced MRI as a poor responsive marker of anti-PD-1/PD-L1 monotherapy for unresectable hepatocellular carcinoma},
  author={Aoki, Tomoko and Nishida, Naoshi and Ueshima, Kazuomi and Morita, Masahiro and Chishina, Hirokazu and Takita, Masahiro and Hagiwara, Satoru and Ida, Hiroshi and Minami, Yasunori and Yamada, Akira and others},
  journal={Liver Cancer},
  volume={10},
  number={6},
  pages={615--628},
  year={2021},
  publisher={S. Karger AG}
}

@article{ye2024gd,
  title={Gd-EOB-DTPA-enhanced MRI proves advantageous in selecting surgical candidates for patients with early-stage hepatocellular carcinoma: An analysis in terms of oncological outcomes},
  author={Ye, Zhiwei and Zhao, Jing and Hu, Dandan and Yang, Zhoutian and Chen, Jinbin and Xu, Li and Zhou, Zhongguo and Chen, Minshan and Zhang, Yaojun},
  journal={iLIVER},
  volume={3},
  number={4},
  pages={100117},
  year={2024},
  publisher={Elsevier}
}

@article{wu2025large,
  title={A large language model improves clinicians’ diagnostic performance in complex critical illness cases},
  author={Wu, Xintong and Huang, Yu and He, Qing},
  journal={Critical Care},
  volume={29},
  number={1},
  pages={230},
  year={2025},
  publisher={Springer}
}

@article{team2025kimi,
  title={Kimi k2: Open agentic intelligence},
  author={Team, Kimi and Bai, Yifan and Bao, Yiping and Chen, Guanduo and Chen, Jiahao and Chen, Ningxin and Chen, Ruijue and Chen, Yanru and Chen, Yuankun and Chen, Yutian and others},
  journal={arXiv preprint arXiv:2507.20534},
  year={2025}
}

@article{yang2025qwen3,
  title={Qwen3 technical report},
  author={Yang, An and Li, Anfeng and Yang, Baosong and Zhang, Beichen and Hui, Binyuan and Zheng, Bo and Yu, Bowen and Gao, Chang and Huang, Chengen and Lv, Chenxu and others},
  journal={arXiv preprint arXiv:2505.09388},
  year={2025}
}

@article{seed2025seed1,
  title={Seed1. 5-thinking: Advancing superb reasoning models with reinforcement learning},
  author={Seed, ByteDance and Chen, Jiaze and Fan, Tiantian and Liu, Xin and Liu, Lingjun and Lin, Zhiqi and Wang, Mingxuan and Wang, Chengyi and Wei, Xiangpeng and Xu, Wenyuan and others},
  journal={arXiv preprint arXiv:2504.13914},
  year={2025}
}

@article{gaber2025evaluating,
  title={Evaluating large language model workflows in clinical decision support for triage and referral and diagnosis},
  author={Gaber, Farieda and Shaik, Maqsood and Allega, Fabio and Bilecz, Agnes Julia and Busch, Felix and Goon, Kelsey and Franke, Vedran and Akalin, Altuna},
  journal={npj Digital Medicine},
  volume={8},
  number={1},
  pages={263},
  year={2025},
  publisher={Nature Publishing Group UK London}
}

@misc{marrocchio2025will,
  title={Will Generative Large Language Models Become Radiologists’ Invaluable Allies?},
  author={Marrocchio, Cristina and Sverzellati, Nicola},
  journal={Radiology},
  volume={315},
  number={2},
  pages={e251259},
  year={2025},
  publisher={Radiological Society of North America}
}

@article{sandmann2025benchmark,
  title={Benchmark evaluation of DeepSeek large language models in clinical decision-making},
  author={Sandmann, Sarah and Hegselmann, Stefan and Fujarski, Michael and Bickmann, Lucas and Wild, Benjamin and Eils, Roland and Varghese, Julian},
  journal={Nature Medicine},
  pages={1--1},
  year={2025},
  publisher={Nature Publishing Group US New York}
}

@article{tordjman2025comparative,
  title={Comparative benchmarking of the DeepSeek large language model on medical tasks and clinical reasoning},
  author={Tordjman, Mickael and Liu, Zelong and Yuce, Murat and Fauveau, Valentin and Mei, Yunhao and Hadjadj, Jerome and Bolger, Ian and Almansour, Haidara and Horst, Carolyn and Parihar, Ashwin Singh and others},
  journal={Nature medicine},
  pages={1--1},
  year={2025},
  publisher={Nature Publishing Group US New York}
}

@article{doi:10.1148/rg.2017170047,
author = {Folio, Les and Machado, Laura and Dwyer, Andrew},
year = {2018},
month = {03},
pages = {462-482},
title = {Multimedia-enhanced Radiology Reports: Concept, Components, and Challenges},
volume = {38},
journal = {RadioGraphics},
doi = {10.1148/rg.2017170047}
}

@article{wang2025credibility,
  title={Credibility assessment of a mechanistic model of atherosclerosis to predict cardiovascular outcomes under lipid-lowering therapy},
  author={Wang, Yishu and Courcelles, Eulalie and Peyronnet, Emmanuel and Porte, Sol{\`e}ne and Diatchenko, Aliz{\'e}e and Jacob, Evgueni and Angoulvant, Denis and Amarenco, Pierre and Boccara, Franck and Cariou, Bertrand and others},
  journal={npj Digital Medicine},
  volume={8},
  number={1},
  pages={171},
  year={2025},
  publisher={Nature Publishing Group UK London}
}

@article{tu2025towards,
  title={Towards conversational diagnostic artificial intelligence},
  author={Tu, Tao and Schaekermann, Mike and Palepu, Anil and Saab, Khaled and Freyberg, Jan and Tanno, Ryutaro and Wang, Amy and Li, Brenna and Amin, Mohamed and Cheng, Yong and others},
  journal={Nature},
  pages={1--9},
  year={2025},
  publisher={Nature Publishing Group UK London}
}

@article{kim2025optimizing,
  title={Optimizing large language models in radiology and mitigating pitfalls: prompt engineering and fine-tuning},
  author={Kim, Theodore Taehoon and Makutonin, Michael and Sirous, Reza and Javan, Ramin},
  journal={RadioGraphics},
  volume={45},
  number={4},
  pages={e240073},
  year={2025},
  publisher={Radiological Society of North America}
}

@article{fink2023potential,
  title={Potential of ChatGPT and GPT-4 for data mining of free-text CT reports on lung cancer},
  author={Fink, Matthias A and Bischoff, Arved and Fink, Christoph A and Moll, Martin and Kroschke, Jonas and Dulz, Luca and Heu{\ss}el, Claus Peter and Kauczor, Hans-Ulrich and Weber, Tim F},
  journal={Radiology},
  volume={308},
  number={3},
  pages={e231362},
  year={2023},
  publisher={Radiological Society of North America}
}

@inproceedings{zhou2021visual,
  title={Visual-textual attentive semantic consistency for medical report generation},
  author={Zhou, Yi and Huang, Lei and Zhou, Tao and Fu, Huazhu and Shao, Ling},
  booktitle={Proceedings of the IEEE/CVF International Conference on Computer Vision},
  pages={3985--3994},
  year={2021}
}

@article{chen2025visual,
  title={Visual-linguistic Diagnostic Semantic Enhancement for medical report generation},
  author={Chen, Jiahong and Huang, Guoheng and Yuan, Xiaochen and Zhong, Guo and Tan, Zhe and Pun, Chi-Man and Yang, Qi},
  journal={Journal of Biomedical Informatics},
  volume={161},
  pages={104764},
  year={2025},
  publisher={Elsevier}
}

@article{chou2025autocodebench,
  title={Autocodebench: Large language models are automatic code benchmark generators},
  author={Chou, Jason and Liu, Ao and Deng, Yuchi and Zeng, Zhiying and Zhang, Tao and Zhu, Haotian and Cai, Jianwei and Mao, Yue and Zhang, Chenchen and Tan, Lingyun and others},
  journal={arXiv preprint arXiv:2508.09101},
  year={2025}
}

@article{zeng2025deepseek,
  title={DeepSeek’s “Low-Cost” Adoption Across China’s Hospital Systems: Too Fast, Too Soon?},
  author={Zeng, Dian and Qin, Yiming and Sheng, Bin and Wong, Tien Yin},
  journal={Jama},
  volume={333},
  number={21},
  pages={1866--1869},
  year={2025},
  publisher={American Medical Association}
}

@article{bhayana2024large,
  title={Large language models for automated synoptic reports and resectability categorization in pancreatic cancer},
  author={Bhayana, Rajesh and Nanda, Bipin and Dehkharghanian, Taher and Deng, Yangqing and Bhambra, Nishaant and Elias, Gavin and Datta, Daksh and Kambadakone, Avinash and Shwaartz, Chaya G and Moulton, Carol-Anne and others},
  journal={Radiology},
  volume={311},
  number={3},
  pages={e233117},
  year={2024},
  publisher={Radiological Society of North America}
}

@article{savage2025radsearch,
  title={RadSearch, a semantic search model for accurate radiology report retrieval with large language model integration},
  author={Savage, Cody H and Chaudhari, Gunvant and Smith, Andrew D and Sohn, Jae Ho},
  journal={Radiology},
  volume={315},
  number={1},
  pages={e240686},
  year={2025},
  publisher={Radiological Society of North America}
}

@misc{yasaka2025new,
  title={A new step forward in the extraction of appropriate radiology reports},
  author={Yasaka, Koichiro and Abe, Osamu},
  journal={Radiology},
  volume={315},
  number={1},
  pages={e250867},
  year={2025},
  publisher={Radiological Society of North America}
}

@article{lucas2025multisequence,
  title={Multisequence 3-T image synthesis from 64-mT low-field-strength MRI using generative adversarial networks in multiple sclerosis},
  author={Lucas, Alfredo and Arnold, Thomas Campbell and Okar, Serhat V and Vadali, Chetan and Kawatra, Karan D and Ren, Zheng and Cao, Quy and Shinohara, Russell T and Schindler, Matthew K and Davis, Kathryn A and others},
  journal={Radiology},
  volume={315},
  number={1},
  pages={e233529},
  year={2025},
  publisher={Radiological Society of North America}
}

@article{zhang2025igniting,
  title={Igniting language intelligence: The hitchhiker’s guide from chain-of-thought reasoning to language agents},
  author={Zhang, Zhuosheng and Yao, Yao and Zhang, Aston and Tang, Xiangru and Ma, Xinbei and He, Zhiwei and Wang, Yiming and Gerstein, Mark and Wang, Rui and Liu, Gongshen and others},
  journal={ACM Computing Surveys},
  volume={57},
  number={8},
  pages={1--39},
  year={2025},
  publisher={ACM New York, NY}
}

@inproceedings{lyu2023faithful,
  title={Faithful chain-of-thought reasoning},
  author={Lyu, Qing and Havaldar, Shreya and Stein, Adam and Zhang, Li and Rao, Delip and Wong, Eric and Apidianaki, Marianna and Callison-Burch, Chris},
  booktitle={The 13th International Joint Conference on Natural Language Processing and the 3rd Conference of the Asia-Pacific Chapter of the Association for Computational Linguistics (IJCNLP-AACL 2023)},
  year={2023}
}

@inproceedings{zhao2025insights,
  title={Insights into deepseek-v3: Scaling challenges and reflections on hardware for ai architectures},
  author={Zhao, Chenggang and Deng, Chengqi and Ruan, Chong and Dai, Damai and Gao, Huazuo and Li, Jiashi and Zhang, Liyue and Huang, Panpan and Zhou, Shangyan and Ma, Shirong and others},
  booktitle={Proceedings of the 52nd Annual International Symposium on Computer Architecture},
  pages={1731--1745},
  year={2025}
}

@article{chan2025deepseek,
  title={DeepSeek-R1 and GPT-4 are comparable in a complex diagnostic challenge: a historical control study},
  author={Chan, Lining and Xu, Xinjie and Lv, Kaiyang},
  journal={International Journal of Surgery},
  volume={111},
  number={6},
  pages={4056--4059},
  year={2025},
  publisher={LWW}
}
\end{document}